\documentclass[a4paper,twoside]{article}

\usepackage{float}
\floatstyle{plaintop}
\restylefloat{table}
\usepackage{placeins}
\restylefloat{table}
\usepackage{epsfig}
\usepackage{subcaption}
\usepackage{calc}
\usepackage{amssymb}
\usepackage{amstext}
\usepackage{amsmath}
\usepackage{amsthm}
\usepackage{multicol}
\usepackage{pslatex}
\usepackage{apalike}
\usepackage{algorithm2e}
\usepackage[bottom]{footmisc}
\usepackage[pagebackref,breaklinks,colorlinks]{hyperref}

\usepackage{SCITEPRESS}     

\begin{document}

\title{Do Object Detection Localization Errors Affect\\Human Performance and Trust?}

\author{\authorname{Sven de Witte, Ombretta Strafforello and Jan van Gemert}
\affiliation{Computer Visions lab, Delft University of Technology, Delft, the Netherlands}
}



\abstract{Bounding boxes are often used to communicate automatic object detection results to humans, aiding humans in a multitude of tasks. We investigate the relationship between bounding box localization errors and human task performance. We use observer performance studies on a visual multi-object counting task to measure both human trust and performance with different levels of bounding box accuracy. The results show that localization errors have no significant impact on human accuracy or trust in the system. Recall and precision errors impact both human performance and trust, suggesting that optimizing algorithms based on the F1 score is more beneficial in human-computer tasks. Lastly, the paper offers an improvement on bounding boxes in multi-object counting tasks with center dots, showing improved performance and better resilience to localization inaccuracy.}

\onecolumn \maketitle \normalsize \setcounter{footnote}{0} \vfill

\section{INTRODUCTION}


Automatic object detectors are used to localize and classify  objects appearing in images and videos. These algorithms have application in a number of fields, including autonomous driving, surveillance, medical imaging, augmented reality, robotics and visual inspection. In this work, we are interested in the applications that involve humans as end users of object detectors. Important examples are anomaly detection in surveillance footage and examination of medical images. 
A common approach of showing the outcome of a object detection system to a human is with bounding boxes. Bounding boxes are rectangular boxes drawn around each object of interest. Object detectors are trained to predict bounding boxes that closely match "ground truth" bounding boxes drawn by humans. 

The quality of object detections is assessed using standard evaluation methods that do not consider the detectors' intended application.
Often, the evaluation is achieved by means of the mean average precision (mAP), a metric that combines object classification and localization accuracy. In particular, an object is considered accurately localized if there is sufficient overlap between the bounding box predicted by the algorithm and the ground truth. The overlap is calculated using intersection-over-union (IoU). Object detections with IoU greater than 0.5 or 0.75 and correct classification are considered acceptable \cite{commonCOCO,everingham2010pascal}. 



\begin{figure}
\includegraphics[width=0.8\linewidth]{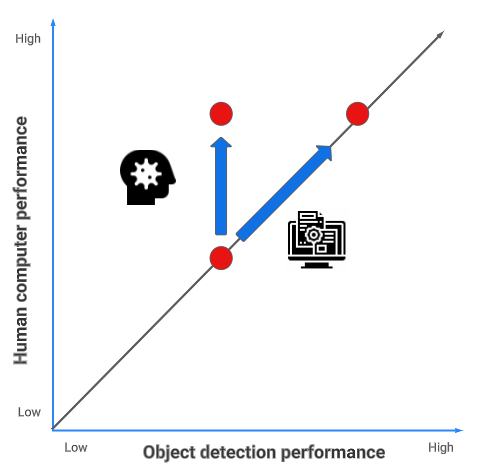}
\caption{Illustration visualizing improvement of human performance in human computer task. A design choice focused on the human in the system could improve performance without need of object detector improvement.}
\label{fig:figure1}
\end{figure}

However, relying on IoU (and therefore, on mAP) can be misleading as IoU is highly affected by small annotation errors present in current datasets \cite{Murrugarra-Llerena_2022_CVPR}.
In addition, previous work shows that the IoU does not consistently align with users preference \cite{strafforello2022humans}. 
In this work, we explore how the performance of the end human users is affected by the object detectors localization errors. We do this using observer performance studies on a practical inspection task including multi-object counting task. In addition, we measure how the human trust towards the object detection system varies as we vary the object detections localization quality. Our study reveals that using center dots instead of boxes not only improves the efficiency of the human in the process, but also increases the overall resilience to inaccurate localization. Generating a performance increase without the need for increased object detector performance as illustrated in Figure \ref{fig:figure1}.




This paper contributes by (1) Designing a study to test human accuracy and performance on a simple object counting task. (2) Showing the relation between object detection accuracy and human performance. (3) Showing the relation between object detection accuracy and human trust. (4) Lastly, showing how simple design choices can increase user performance.


\section{RELATED WORK}

Research on object detectors aims to enhance the localization and classification performance on a variety of images and datasets, accelerate the inference speed and reduce the computational requirements. Current models comprise two-stage models \cite{cai2018cascade,girshick2015fast,girshick2015region,ren2015faster}, single stage approaches \cite{lin2017focal,ssd,yolo,redmon2018yolov3}, pointwise/anchorless methods \cite{duan2019centernet,law2018cornernet,zhou2019bottom}, transformers-based detectors \cite{li2022exploring,beal2020transformerbased,Carion_2020,Dai_2021_CVPR,zhu2020deformable} and, more recently, diffusion models \cite{chen2023diffusiondet}.
%

Although the latest state-of-the-art models have achieved competitive performance on standard benchmarks, progress in object detection is often carried out overlooking what the intended application of object detectors is. In fact, \cite{strafforello2022humans} showed that when object detections are meant to be shown to humans, standard evaluation metrics are unreliable.
In this work, our objective is to assess the impact of object detections localization accuracy when they are intended for applications involving human users. In this regard, we conduct an observer performance study to compare object detectors localization accuracy with users performance.  

Using observer performance study to analyze and improve the workflow between humans and computer vision systems is not new. \cite{valueOfObserverPerf} deployed observer performance studies to assess assistive imaging techniques in radiology.  Similarly \cite{adsOPS} looked at the impact computer aided detection for radiologist using 3D MR imaging.

When looking at the mixed field of psychology and computer science, \textit{trust} is an important factor to consider. Extensive research has been conducted on effectively measuring human trust in computer systems, yet it remains a challenging task.
%
Multiple studies examining trust measures and methodologies are available \cite{MeasuresOfTrustInAutomation,reviewOfTrustMeasures,InvestigatComponentMeasures}. However, it has been demonstrated that many studies employ trust measures that are either inadequately validated or specifically designed for a particular use case \cite{MeasuresOfTrustInAutomation}.
%
%
%
In this study, we use the multi-dimensional scale proposed by \cite{TrustSurvey} to assess user trust in human-computer interactions. 

\section{METHOD}

Our goal is to investigate the relation between object detection localization accuracy and human performance and trust. To do this, we conduct an observer performance study. We design a task where participants of the study have to count the number of aquatic creatures in images.


In the experiments, participants are shown images containing bounding boxes and center dots with varying localization accuracy. The answers and response times per image are recorded. At the end of the task, the participants are asked to fill in a survey related to the trust in the system that generated the bounding boxes. 
Participants are informed that the same system could potentially be utilized to monitor the growth and well-being of creatures in the aquarium or for an automated feeding system, where errors might result in incorrect meal sizes.
The survey consists of 12 question about the perceived risk, benevolence, competence and reciprocity. Respondents answer through a Likert scale, ranging from 1 (strongly disagree) to 5 (strongly agree).

The errors introduced are using 2 different measures. The IoU that gives a bounding box a score based on the overlap of the placed box compared to the ground truth. The Shifted boxes and Shifted Dots are shifted into a random direction from the ground truth to create an IoU of 0.5. The other measure we use is the F1 score. The F1 score is calculated using the precision and recall. Precision measures how accurate a object detector is looking at the amount of true positive boxes placed compared to the total amount of boxes placed. This score is varied by placing additional boxes. Recall measures the ability of the object detector to find all the objects of interest calculated by dividing the number of true positive boxes by the total number of boxes that the ground truth has. A lower recall can be achieved by removing some of the ground truth boxes.
%
The object detections proposed in our study are generated by manipulating ground truth bounding boxes. We do not make use of real object detectors to ensure full control of the object detections localization error. 
Our study participants are recruited through Amazon Mechanical Turk \cite{mturk}. 

In total we perform 7 experiments, each containing a set of 30 images divided into 3 smaller tasks of 10 images. This choice was made to reduce the workload per participant, increasing their cooperation and reducing the chance of noise in our results. Every task is performed by 10 to 12 participants, resulting in approximately 36 participants per experiment following the recommendations of~\cite{HowManyCrowdsource}.


Within this study, we measure the participants error by calculating the absolute difference between the real number of aquatic creatures in an image and the provided answer. We also calculate the agreement between the participants using the Krippendorff's Alpha \cite{krippendorff2010vols} and use it to assesses the participants' reliability. 
%
To test the significance of our findings we use the T-test with Bonferroni correction. This study involves 7 different groups of data resulting in significance threshold $\alpha = \frac{0.05}{7} = 0.0071$. 


\section{Experiments}

\subsection{Dataset}
\begin{figure*}[t]
\begin{tabular}{@{}cccccc@{}}
Perfect box & Shifted box & Perfect Dot & Shifted Dot & False positive & False negative \\
\includegraphics[width=0.15\linewidth]{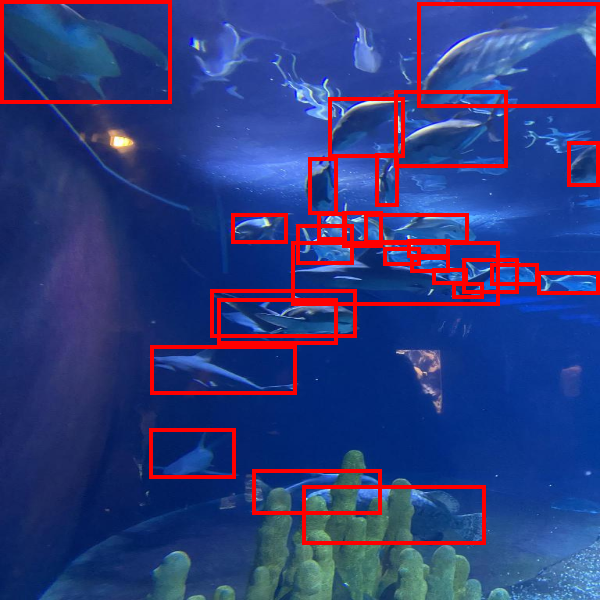} &
\includegraphics[width=0.15\linewidth]{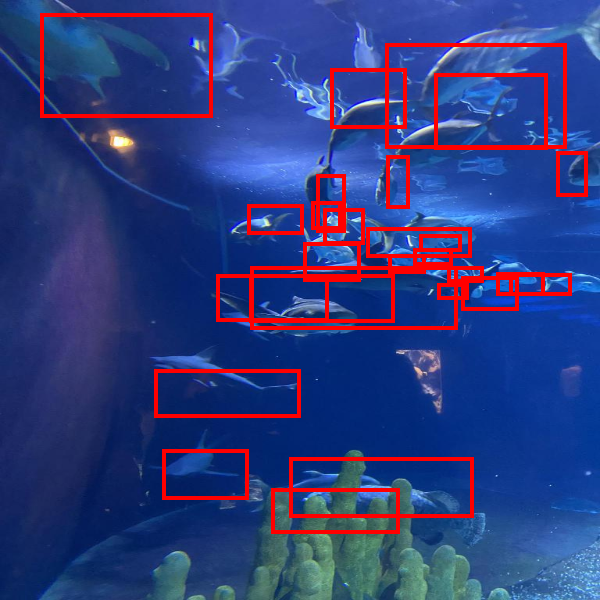} &
\includegraphics[width=0.15\linewidth]{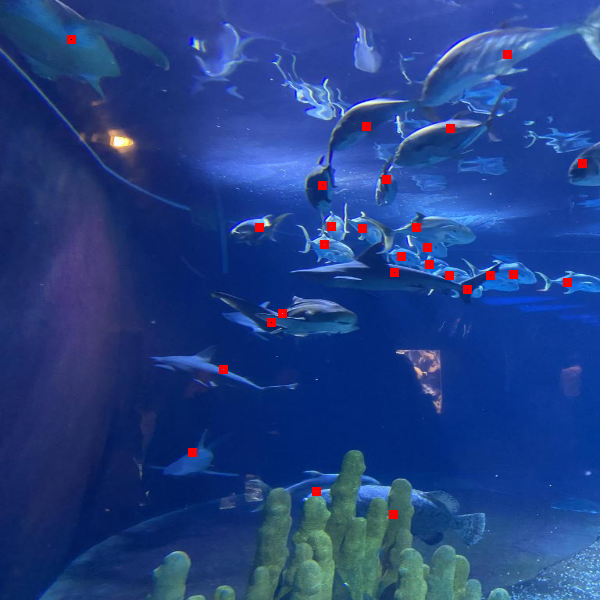} &
\includegraphics[width=0.15\linewidth]{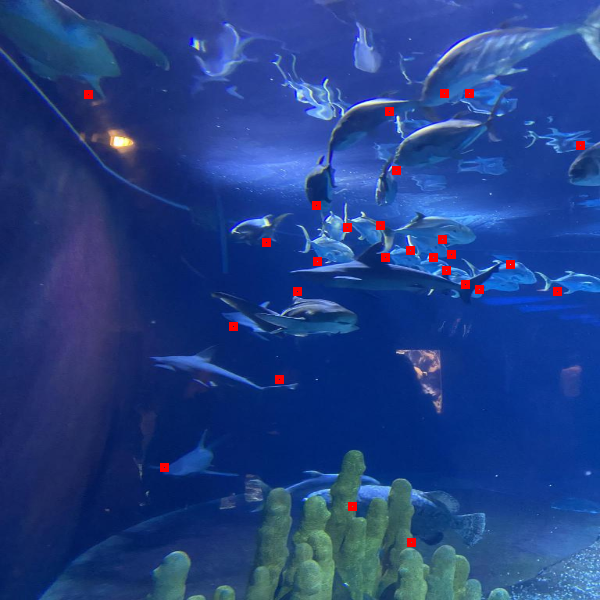} &
\includegraphics[width=0.15\linewidth]{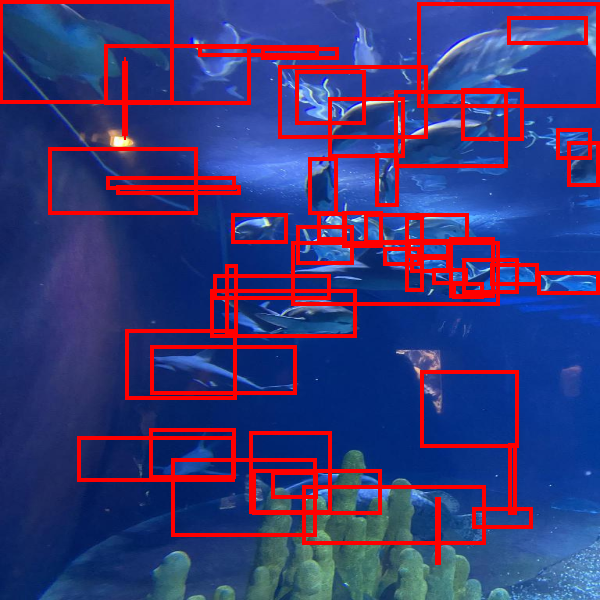} &
\includegraphics[width=0.15\linewidth]{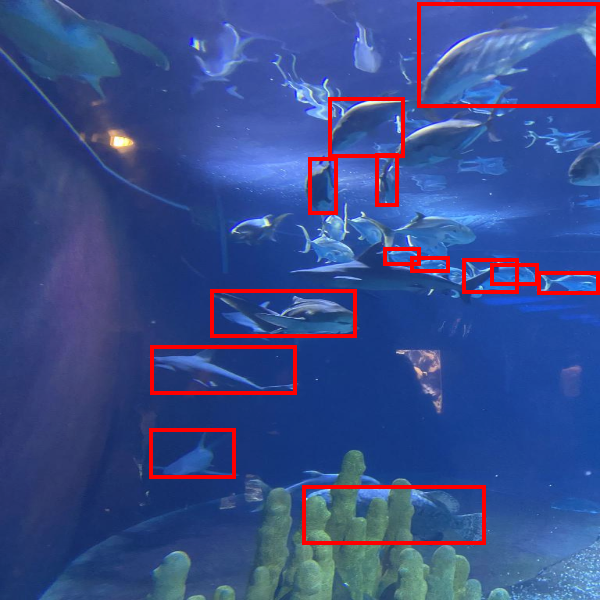} \\
\includegraphics[width=0.15\linewidth]{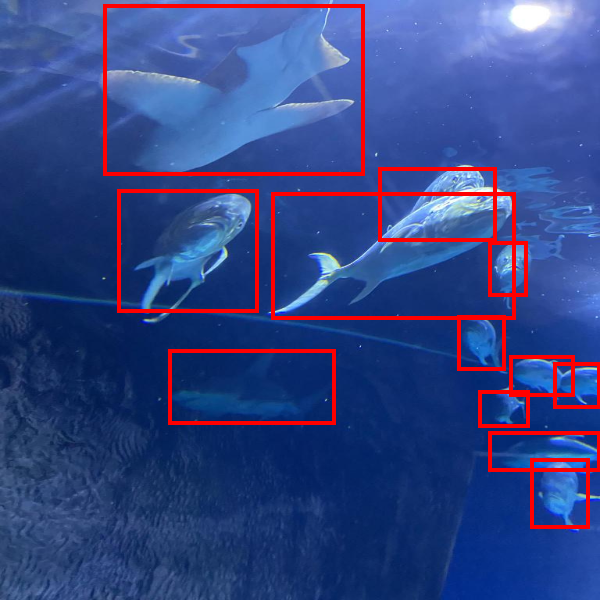} &
\includegraphics[width=0.15\linewidth]{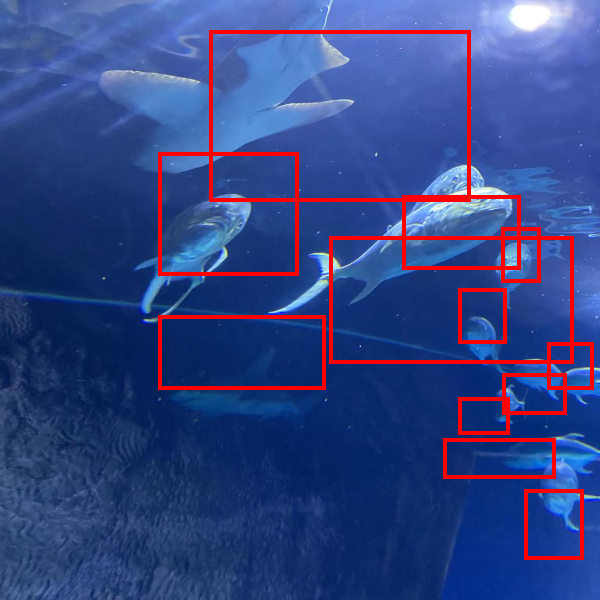} &
\includegraphics[width=0.15\linewidth]{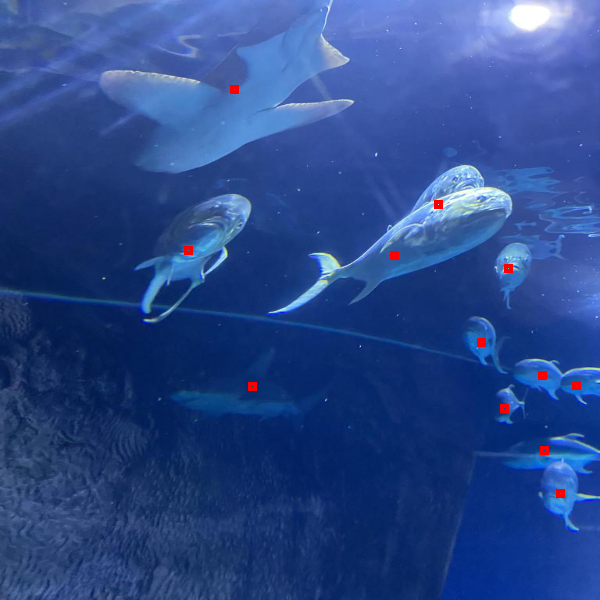} &
\includegraphics[width=0.15\linewidth]{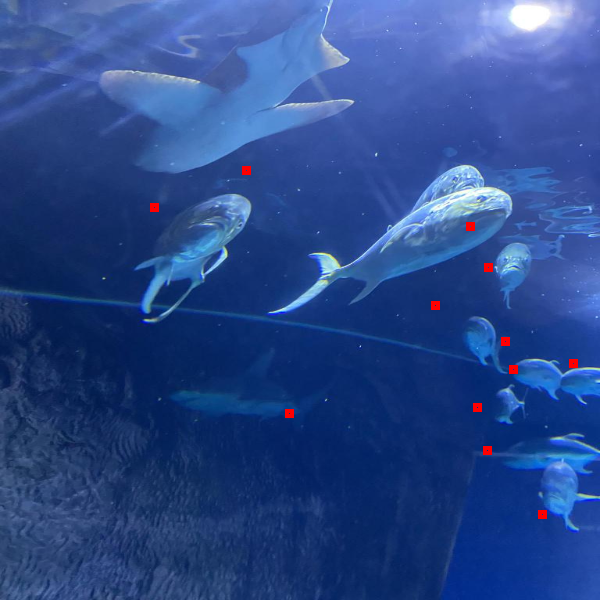} &
\includegraphics[width=0.15\linewidth]{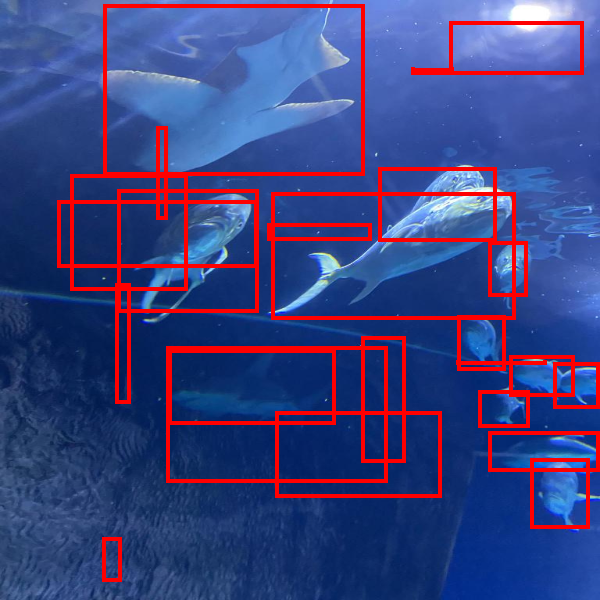} &
\includegraphics[width=0.15\linewidth]{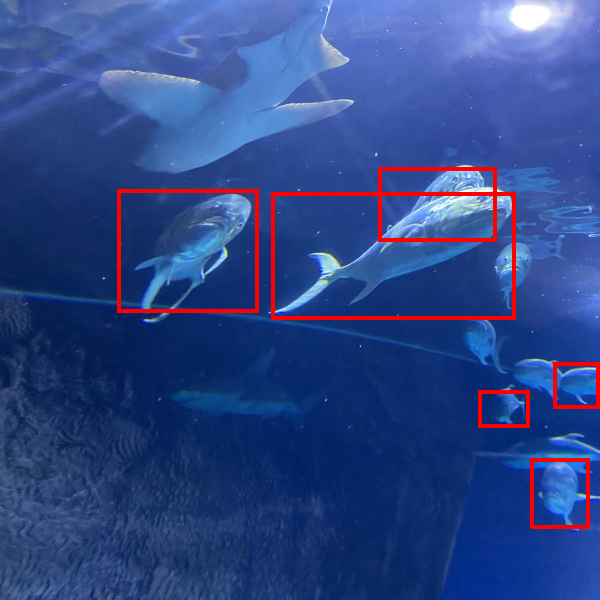}
\\
\includegraphics[width=0.15\linewidth]{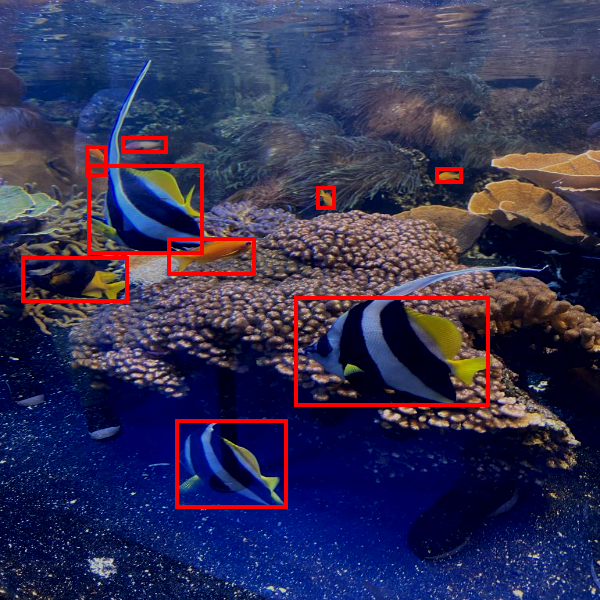} &
\includegraphics[width=0.15\linewidth]{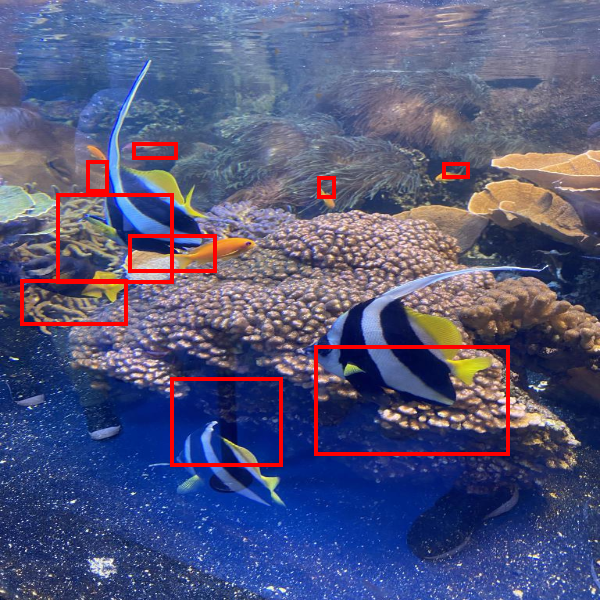} &
\includegraphics[width=0.15\linewidth]{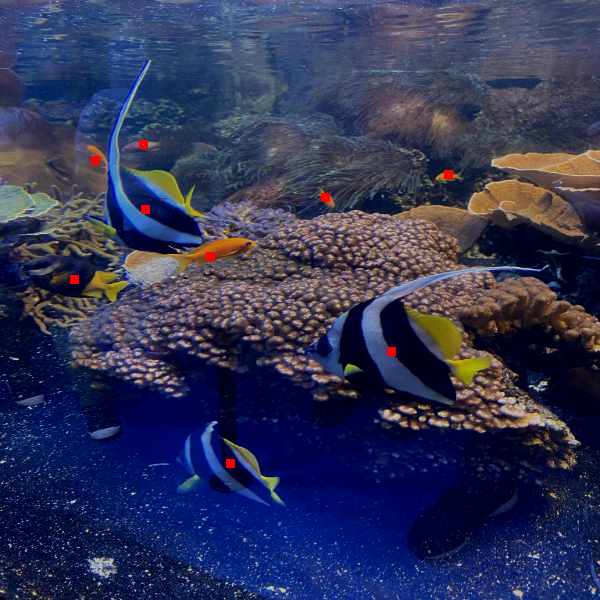} &
\includegraphics[width=0.15\linewidth]{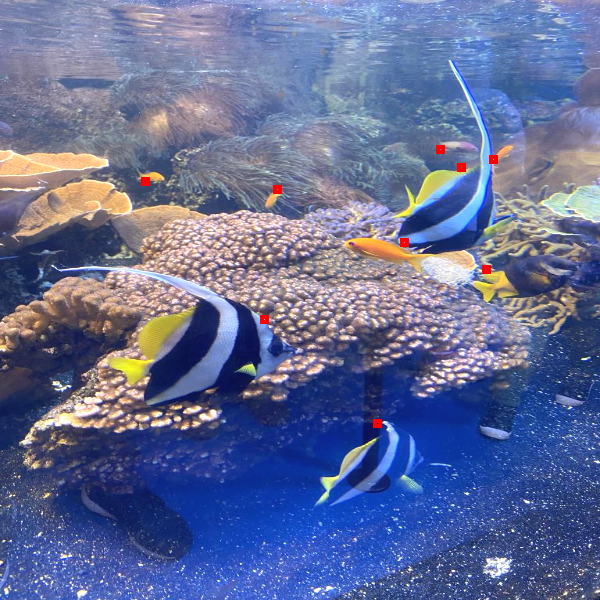} &
\includegraphics[width=0.15\linewidth]{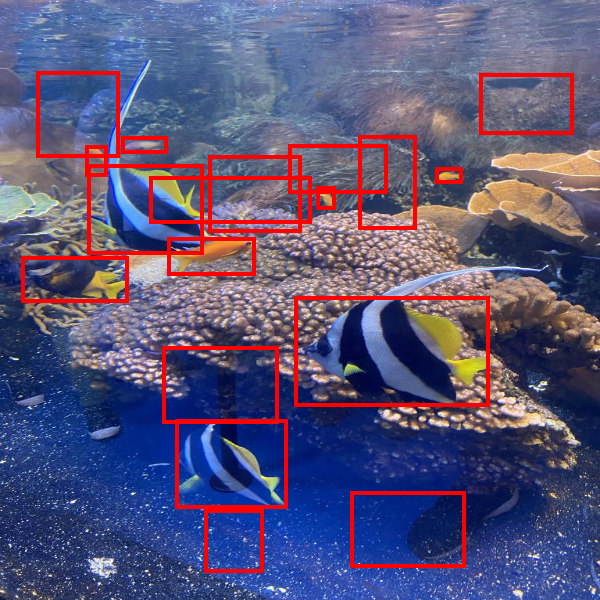} &
\includegraphics[width=0.15\linewidth]{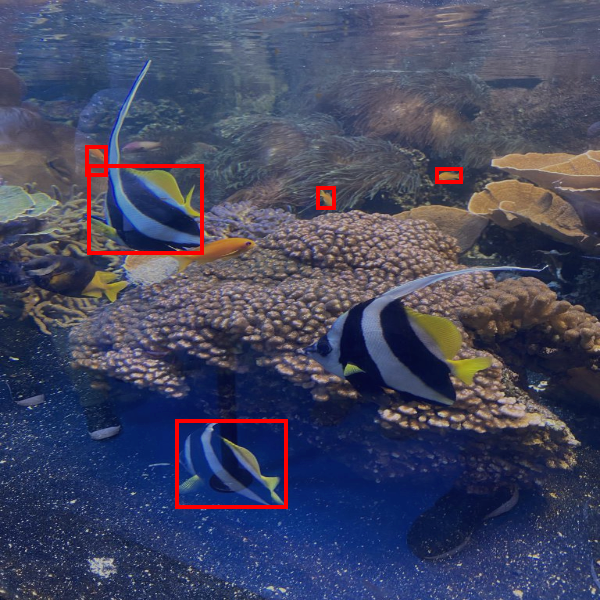}
\end{tabular}
\caption{Example of images used within the study. Going from left to right accurate bounding boxes (\textit{Perfect box}), bounding boxes with 0.5 IoU (\textit{Shifted box}), accurate Dots (\textit{Perfect Dot} ), Inaccurate dots (\textit{Shifted Dot}), additional boxes(\textit{False positive}) and missing boxes(\textit{False negative}).}
\label{fig:survey_images}
\end{figure*}

All images used in this study come from the Aquarium combined data set from Roboflow\cite{dwyer2022,Aquarium}. 
The dataset contains images with multiple aquatic creatures. The images were labeled by the Roboflow team with help of SageMaker Ground Truth. From this dataset the 30 most suitable images for the experiments were hand picked. To be suitable for the experiment an image has to have multiple creatures in the image, the creatures must be in the water (images containing puffins outside the water could lead to confusion), the image must be clear and the creatures in the image have to be big enough to be able to recognize as a aquatic creature. The bounding box information of some of the images were updated after finding additional unmarked fish that were in the image or reflections of fish that were marked by the original team. 
All bounding boxes and dots are drawn is a bright red color. As red contrasted the best with the dark and blue tones that were most present in the images. Examples of the different kind of images are visualized in the grid in Figure \ref{fig:survey_images}.


\subsection{Pilot Testing}

Prior to conducting the main survey on Amazon Mechanical Turk, we conducted pilot surveys in the MTurk sandbox to evaluate the clarity of images and accuracy of bounding box surveys. The purpose of this pilot study was to gather feedback on the survey and assess the task's difficulty level. If the task was too easy, participants would likely achieve close to 100\% accuracy with short response times, potentially diminishing the performance differences between various experiments. On the other hand, if the task was too difficult, overall performance across all experiments would likely be poor. The accuracy results from the two pilot surveys indicated that the task's difficulty level was appropriate for testing the hypotheses.

Additionally, the pilot studies served as a means to estimate the average time required to complete the survey. This information was essential for ensuring that participants received appropriate compensation for their time investment in the task. Moreover, the pilot surveys helped identify any errors in the images and any ambiguities in the task or images that could lead to divergent subjective interpretations of the correct answers. Such discrepancies would undermine or interfere with the hypotheses being tested.

\subsection{Clean Image Baseline}
The first experiment aimed to determine the task difficulty and establish a baseline for overall performance. Clean images, consisting of the 30 images used in this study without additional information such as bounding boxes, were employed. These 30 images were divided into three counting tasks, each containing 10 images. 

Analyzing the results of the individual counting tasks gives an average error of 3.64, 3.68, and 3.88. These results indicate that the task difficulty remains relatively consistent across all the counting task. 
In this study, the Krippendorff's Alpha value 
was found to be 0.76, indicating a substantial level of agreement among participants regarding the correct answers. A low score would indicate disagreement or unreliability in the data, which would mean that the images or task are ambiguous or data is labeled wrong. A score of 0.76 suggest that the data and task are clear and we can use the outcome.
\newline\newline
The overall average error for the clean image task was 3.7, with a standard deviation of 3.6 (see Table \ref{shiftboxstats}). The standard deviation was calculated based on the absolute errors for all questions, without distinguishing between errors above or below the correct answer.
These findings serve as a valuable baseline for the subsequent experiments, providing insights into the task difficulty and establishing a reference Dot for performance comparison.

\subsection{Perfect Bounding Boxes}
The objective of the second experiment is to assess whether adding a correct bounding boxes to the images aids the participants' performance.  
We use the same experimental setup as for first experiment. 

As presented in Table \ref{shiftboxstats}, the average error and standard deviation of errors in this experiment were significantly lower than those observed in the baseline experiment. As anticipated, participants exhibited improved performance in accurately counting the fish when provided with the correct bounding box information. Furthermore, participants spent less time on the images in this experiment. However, we found no statistical significant difference in response times. 
This implies that participants spent roughly the same amount of time on images both with and without bounding boxes, despite their enhanced performance when the bounding boxes were available.

These findings suggest that the inclusion of correct bounding boxes in the images led to a significant reduction in errors and improved participant performance in counting the fish. Moreover, the response times remained comparable, indicating that participants efficiently processed the images with or without the presence of bounding boxes, albeit with superior results when aided by the bounding box information.

\begin{figure}[!ht]
  \centering
   {\epsfig{file = 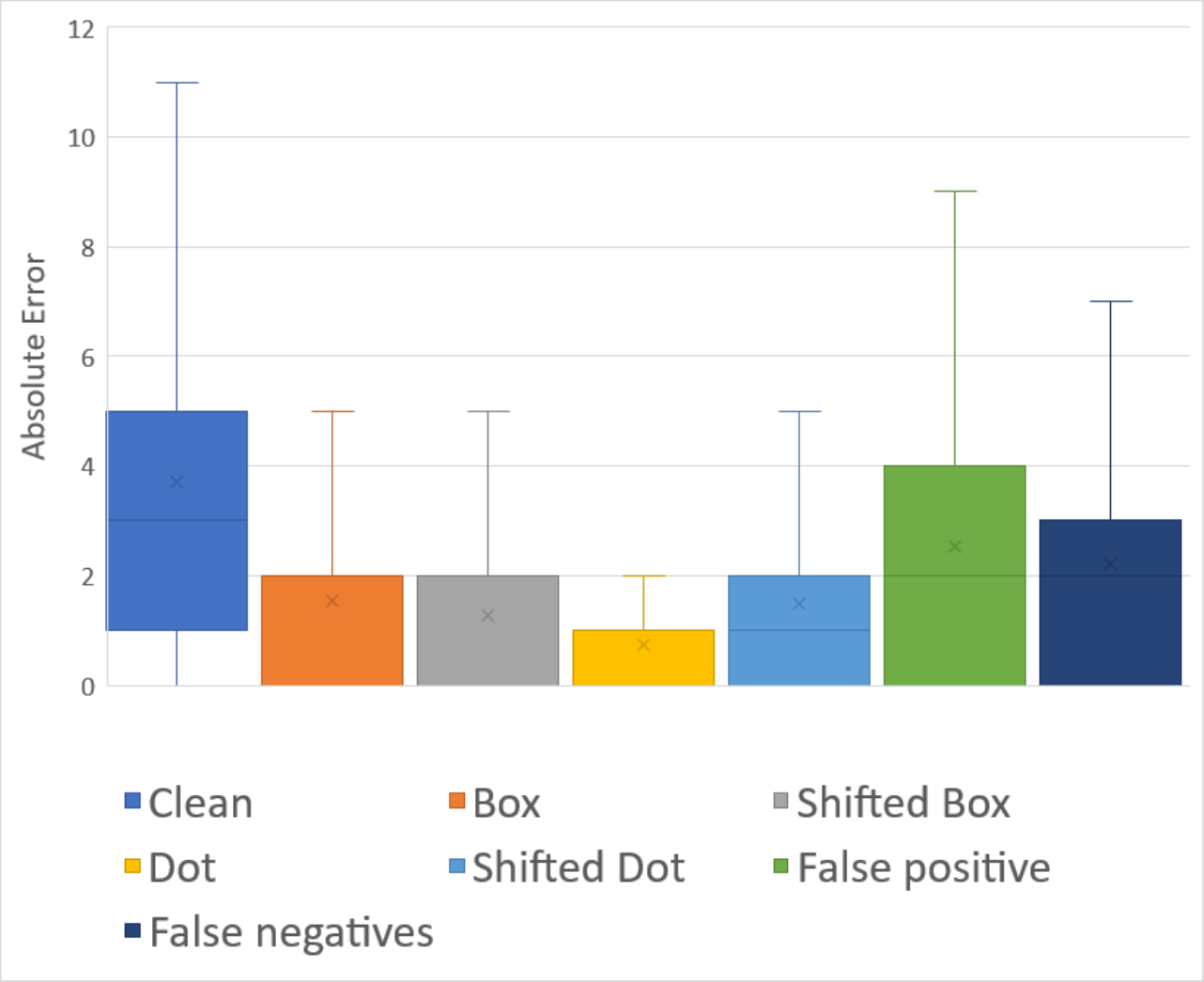, width = 7.5cm}}
  \caption{Boxplot comparing the absolute error of the different experiments. }
  \label{fig:Errors}
 \end{figure}

 \begin{figure}[!ht]
  \centering
   {\epsfig{file = 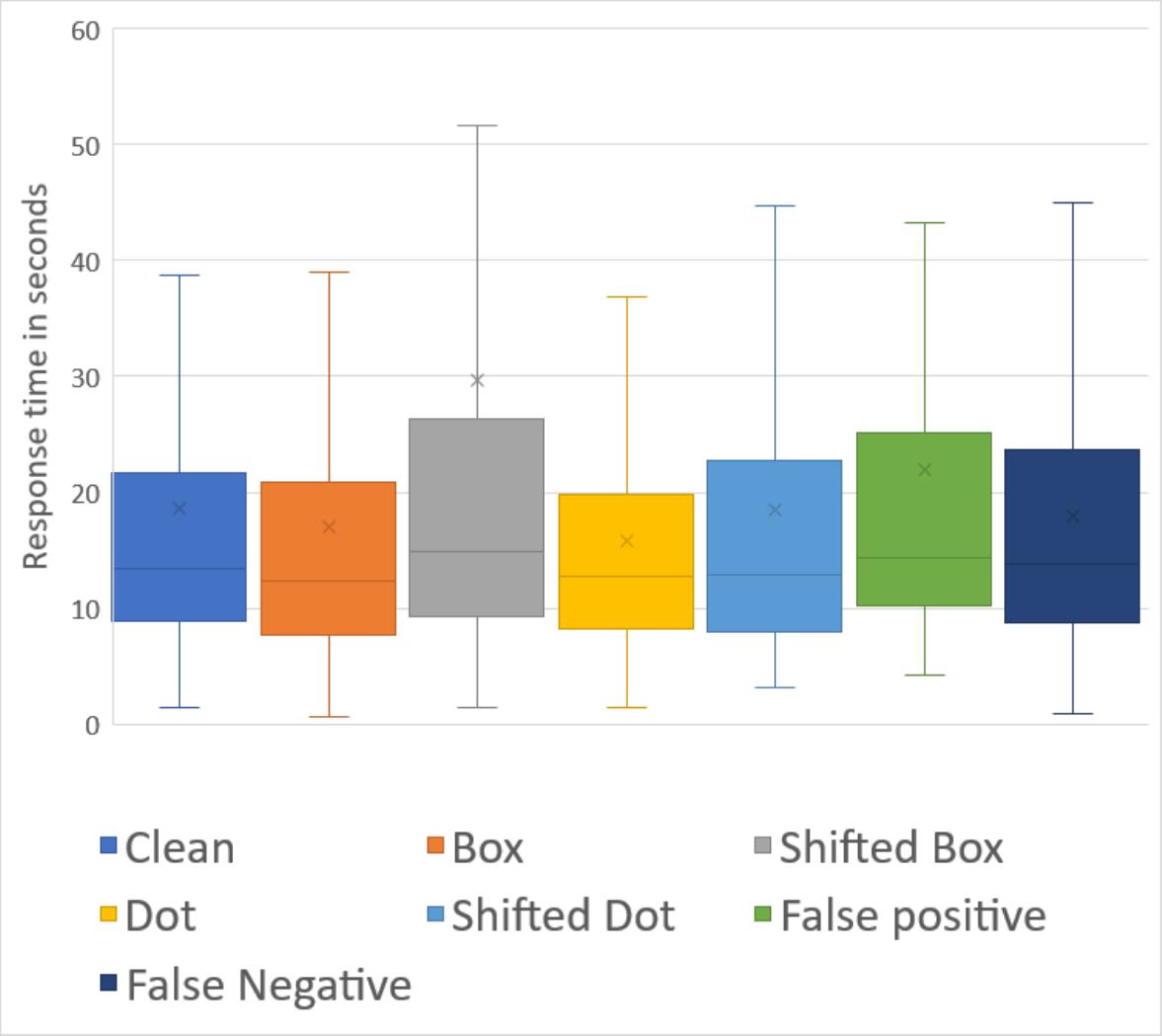, width = 7.5cm}}
  \caption{Boxplot comparing response times within the different experiments.}
  \label{fig:Times}
 \end{figure}


\subsubsection{Shifted Bounding Boxes}\label{sec:shiftedbbox}
The purpose of the third experiment is to evaluate the performance of humans when presented with images containing bounding boxes with lower object detection localization accuracy. In this experiment, the bounding boxes in the images are randomly shifted in direction to create an IoU value of 0.5. The scale of the bounding boxes remains consistent, as variations in bounding box scale could introduce additional factors that might influence the experiment's outcomes. Similarly to the previous experiments, the images were divided into three counting tasks, with a total of 36 participants.

As shown in Table \ref{shiftboxstats}, the average error observed in this experiment does not appear significantly different compared to the experiment with correct bounding boxes. Surprisingly, the participants' performance is even better when confronted with the shifted bounding boxes. Conducting a T-test on the data confirms that there is no significant difference in accuracy between the two types of images. However, a notable difference is observed in the response times for the images with participants taking an average of 12 seconds more of the images containing the shifted boxes. This finding suggests that overall performance on the images is better when participants are provided with correct bounding boxes. The discrepancy in response times can be attributed to participants needing to count more carefully on the images with shifted bounding boxes compared to those with correct bounding boxes. 



\subsection{F1 Errors}
Localization errors are not the only type of errors that object detectors struggle with. A object detector can fail to find the objects of interest or might detect a object where there is none. These are called type II and type I errors or false-positives and false-negatives. The F1 score is used to give the object detector a score based on these errors . 

Two experiments were run to analyze the impact of type I and type II errors on the human performance and trust. The hypotheses for these two experiments were that the overall performance and trust of the humans would be lower than in the case of the correct or shifted boxes. This because the amount of boxes shown does not equal the correct amount of sea creatures within the image. 

\begin{table}[H]
\caption{
Participant mean error and response time when presented with Clean images (\textit{Clean}). accurate bounding boxes (\textit{Perfect box}), bounding boxes with 0.5 IoU (\textit{Shifted box}), accurate Dots (\textit{Perfect Dot} ), Inaccurate dots (\textit{Shifted Dot}), additional boxes(\textit{False positive}) and missing boxes(\textit{False negative}). }
\begin{tabular}{l|ll}
            & Mean error & Mean response time \\ \hline
Clean       & $3.7 \pm 3.6$        & $18.6 s \pm 17.4 s$               \\
Perfect Box &  $1.5 \pm 2.7$        & $17.0s \pm 15.8 s$               \\
Shifted Box & $1.3 \pm 2.0$       &  $29.7 s \pm 7.3 s$             
\\
 Perfect Dot& $0.7 \pm 1.8$&$15.8 s \pm 11.7 s$\\
 Shifted Dot& $1.5 \pm 2.5$&$18.5 s \pm 16.8 s$\\
 False positive& $2.5 \pm 2.7$&$21.9 s \pm 21.8 s$\\
 False negative& $2.2 \pm 2.2$&$17.9 s \pm 12.2 s$\\\end{tabular}
\label{shiftboxstats}
\end{table}

\subsubsection{False-Positives}
In this experiment, the images were altered to have additional boxes on random locations in the image as demonstrated in the 5th column of the image grid in Figure \ref{fig:survey_images}. The bounding box size was in a range around the average box size within the image. The amount of boxes in the image was doubled to create a precision score of 0.5. 

As the data in Table \ref{shiftboxstats} shows, the average error is 2.524. Using a t-test with an $\alpha = 0.0071$ it shows that this difference is significant compared to that of correct bounding boxes. The same goes for the average time spent per image with an average of 21.891 seconds.
This result is important as it shows that false-positive errors impact the performance of the human both in accuracy and speed.

\subsubsection{False-Negatives}
In this experiment, we remove 50\% of the bounding boxes in each image, leading to a decrease of the recall score to 0.5 as demonstrated in the 6th column of figure \ref{fig:survey_images}. This way not all sea creatures are as easy to find as in the experiments with perfect recall, and there are clearly visible mistakes from the object detector.

The data in Table \ref{shiftboxstats} shows an average error that is significantly higher compared to the correct bounding boxes. With an average error of 2.216. This shows that the participants had an overall worse performance when not all of the boxes are present in the image. However, we find no significant difference in the response time. This means that, on average, people spent about the same amount of time on false negative images as they did on the images with the correct bounding boxes.

\begin{table*}[t]
\caption{Result summary of trust scores per experiment out of 100. The data shows no signifact drop in trust with perturbations to the box and dots location, while adding or removing false and true positive bounding boxes does. This suggest to optimize for precision and recall in human computer tasks. }
\label{tab:trustscores}
\begin{tabular}{|l|l|l|l|l|l|l|}
\hline
            & Perfect box & Shifted Box & Perfect Dot & Shifted Dot  & False positive & False negative\\ \hline
Trust score & $63.7 \pm17.5$       & $62.5 \pm 16.0$      & $57.7 \pm 10.0$      & $63.7 \pm 16.4$      & $28.8 \pm 12.9$         & $23.9 \pm 8.7$\\ \hline
\end{tabular}

\end{table*}

\subsection{Perfect Center Dots}
The introduction of center dots is proposed as a potential solution to address challenges associated with bounding boxes such as occlusion, overlapping and clutter. In this experiment, images are modified to include red dots positioned at the center of where the original bounding boxes would be, effectively marking all aquatic creatures with a red dot. The hypothesis states that images with center dots will yield superior human performance and accuracy compared to images with correct bounding boxes. This hypothesis stems from the notion that bounding boxes in a multiple object counting task may introduce additional difficulties for certain images.

As depicted in some of the images in Figure \ref{fig:survey_images} the proximity of multiple bounding boxes can result in the formation of new boxes through overlap. These new boxes could potentially confuse or mislead participants attempting to count the number of aquatic creatures as quickly as possible, leading them to either perceive empty boxes or count additional boxes. To address this issue, the presented approach utilizes center dots, represented by down scaled versions (9 by 9 pixels) of the original bounding boxes. The expectation is that employing center dots will enhance human performance and accuracy while potentially reducing the overall Intersection over Union (IoU) metric.

Table \ref{shiftboxstats} illustrates that the overall task performance using center dots surpasses that of correct bounding boxes. With a mean error of 0.74, the center dot approach achieves significantly lower error rates compared to the 1.53 error associated with correct bounding boxes in this simple task. Utilizing a T-test with a significance threshold $\alpha$ of 0.0071 demonstrates that the difference in performance is statistically significant and not merely a result of random chance. Conversely, no significant difference is observed in response times between the two approaches. This finding is intriguing, as it indicates that participants spent an equivalent amount of time on both types of questions, while accuracy with center dots was higher.

The results indicate that incorporating center dots leads to superior performance and accuracy in the task compared to the use of correct bounding boxes. This outcome holds even though participants spent similar amounts of time on both types of questions



\subsubsection{Shifted Center Dots}
How do participants perform when the location of the center dots is inaccurate? This experiment is similar to the \textit{Shifted Bounding Box} (Section \ref{sec:shiftedbbox}), but in this case, the dots are deliberately shifted to achieve lower accuracy. The images in this experiment include center dots that are randomly displaced in a direction based on where the center of the box would be with a 0.5 IoU when drawing the bounding boxes. We hypothesize that the advantage of over bounding boxes found in the \textit{Perfect Center Dots} experiment will still hold, even with reduced accuracy. This is because there is a greater likelihood that the center dot will still connect to a portion of the object. Additionally, center dots may aid in human perception of a less accurate system. Bounding boxes make it easier for humans to perceive the boundaries of the box, as demonstrated in the study by List and Bins \cite{EvaluatingEvaluator}.

The results indicates that the participants performed with similar accuracy and time compared to the correct bounding boxes. Using a T-test with a significance level of $\alpha$ 0.0071 on the times of the shifted bounding boxes and shifted dots reveals a significant difference. This suggests that the overall performance of the shifted dots, although inferior to that of the center dots, is better than that of the shifted bounding boxes. The aforementioned perceptual factors could also play a role, where the reduced accuracy of the dots might be less noticeable, resulting in people being less cautious compared to their interaction with the shifted bounding boxes.

\subsection{Trust} 
In addition to the participants performance, we measure the participants trust towards the object detector.
%
%
Table \ref{tab:trustscores} shows the collected trust scores, ranging on a scale from 1 to 100. Surprisingly, there is no significant difference in trust when showing participants correct bounding boxes versus boxes with localization errors. 
This might suggest that users solved the counting task by counting the number of bounding box present in the images, while not focusing too much on the exact box location.
\newpage
On the other hand, the trust scores of the experiments with the false negatives and false positives are significantly lower than the \textit{perfect box} baseline. This suggests that showing an incorrect amount of bounding boxes significantly impact the participants perception of the object detection system. 

The trust scores collected in presence of false negatives and false positives align with the diminished performance of participants observed in these two experiments, suggesting that users' self-perception of their own performance plays a pivotal role in determining their trust in the system.

\section{DISCUSSION}

In this work, we show that the human performance in a visual multi-object counting task is improved when introducing an assistive object detection system.
%
%
Secondly, we measure how human performance varies as we control for object detector localization errors.
Interestingly, we find that perturbing the location of the object detections does not degrade the human counting accuracy nor their trust in the system, but only increases the completion time of the task. 
In addition, human performance is improved 
even when the object detections are not perfect, but contain precision and recall errors. 
%
%
On the other hand, we find that the human trust in the object detector system significantly decreases in presence of false positive or false negative detections. Therefore, we conclude that when object detections are meant to be presented to humans, it is more important to optimize the F1-score over the IoU.

Finally, we show that the visualization strategy used to present the object detections has an impact on the human performance. In fact, in our study, using center dots instead of bounding boxes increases the human performance in the object counting task and the resilience to localization errors. This highlights the importance of presenting data in a way that is optimal for the task.

\bibliographystyle{apalike}
\bibliography{main.bib}



\end{document}